\documentclass{article}
\usepackage{ijcai17}

\usepackage{times}

\usepackage{amsmath,graphicx}

\title{Privacy Preserving Face Retrieval in the Cloud for Mobile Users}
\author{Xin Jin$^1$, Shiming Ge$^{2,*}$, Chenggen Song$^1$ \\
$^1$Department of Computer Science and Technology\\
Beijing Electronic Science and Technology Institute, Beijing, 100070, P.R. China\\
$^2$Institute of Information Engineering\\
Chinese Academy of Sciences, Beijing, 100093, P.R. China
\\$^{*}$Corresponding author: geshiming@iie.ac.cn}

\begin{document}

\maketitle

\begin{abstract}
Recently, cloud storage and processing have been widely adopted. Mobile users in one family or one team may automatically backup their photos to the same shared cloud storage space. The powerful face detector trained and provided by a 3rd party may be used to retrieve the photo collection which contains a specific group of persons from the cloud storage server. However, the privacy of the mobile users may be leaked to the cloud server providers. In the meanwhile, the copyright of the face detector should be protected. Thus, in this paper, we propose a protocol of privacy preserving face retrieval in the cloud for mobile users, which protects the user photos and the face detector simultaneously. The cloud server only provides the resources of storage and computing and can not learn anything of the user photos and the face detector. We test our protocol inside several families and classes. The experimental results reveal that our protocol can successfully retrieve the proper photos from the cloud server and protect the user photos and the face detector.
\end{abstract}

\section{Introduction}
In today's mobile Internet era, increasing mobile users backup their photos to the cloud storage servers. Some cloud servers provide face retrieval service, which allows one to retrieve photos that contain a specific person or a group of persons from all his/her photos in the own storage space of the cloud server. 

\begin{figure}
\centering
\includegraphics[height=4.7cm]{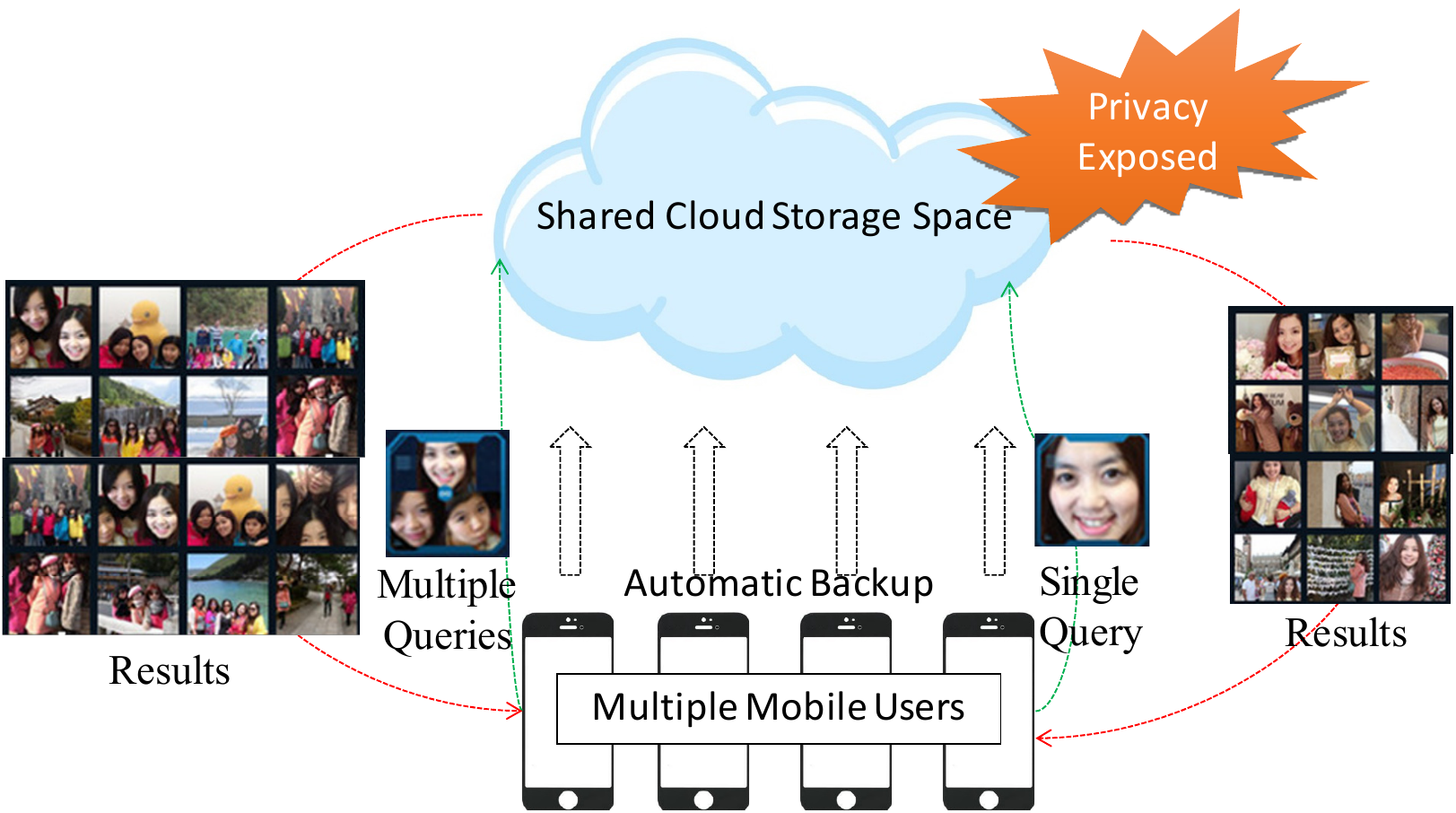}
\caption{The typical scenario. Multiple users backup and share their photos in their mobile phones to a cloud storage. However, the privacy of photos is completely exposed to the cloud.}
\label{fig:scenario}
\end{figure}

Further more, as shown in Figure \ref{fig:scenario}, people in one team or one family may share the same cloud storage space and upload their photos together.  For example, people in one family take photos of each other using their mobile phones for a long time. In the traditional way, they may copy photos in each mobile phone using a cable and manage their photos manually. Nowadays, one can create a cloud storage space and share it to all the family members. All the photos shot by the family members can be stored to the shared cloud storage space automatically by the cloud Apps in their mobile phones. After that, each family member can browse photos using the cloud Apps in a friendly way. Besides, one can retrieve the photos contains one or multiply specific family members using the photo management module of the cloud Apps, as shown in Figure \ref{fig:scenario}.

The above application models are typical scenarios in today's mobile Internet era. However, although the management of group photos is much more convenient, the privacy of the users' photos is completely exposed to the cloud server. The facial features of each member in one family, the relationship between members, and the school of their children, etc. can be learned from the family photos, which can threaten the personal and property security of the family.

In the meanwhile, the face detector used in the face retrieval task may be trained in a large scale of face images annotated by thousands of people. The copyright of the trained parameters of the face detector should also be preserved from the commercial provider's perspective.

Thus, in this paper, we propose a novel protocol to preserve the privacy of the cloud users' photos and the parameters of the commercial face detector simultaneously in such mobile cloud scenarios. The face retrieval problem can be decomposed into \emph{\textbf{face detection}}, \emph{\textbf{face recognition}} and \emph{\textbf{face label matching}}. In the face detection stage, face regions are detected in users' photos with rectangles. In the recognition stage, each detected face is marked by a label of a member in a group. Then a label vector is generated according to the face recognition result for each photo so as to mark who is/are in each photo. The above is the off-line phase. In the on-line phase, a user queries a specific face of one person or faces of a group of person. Then, a label vector is generated for this query and compared to each label vector corresponding to each photo. Photos with the most similar label vector to the query label vector are selected as the retrieval result.

\textbf{Related Work}. The secure face detection method is proposed as Blind Vision \cite{Avidan2006a} for securely evaluating a Viola-Jones type face detector. After that Jin et al. accelerate secure face detector by introducing a random base image representation \cite{JinICME2017}.  A system called Secure Computation of Face Identification (SCiFI) \cite{Osadchy2010} is developed for secure face recognition.  This system use two cryptography tools (homomorphic encryption and oblivious transfer) to implement a privacy preserving computation of the Hamming distance between two binary vectors. Recently, A lot of researchers have addressed the privacy preserving computer vision problems \cite{Shashank2008,Upmanyu2009,Osadchy2010,Sohn2010,Fanti2013,Chun2014,Bost2015,Jin2016,JinWCSP2016}. Most of them leverage the cryptography tools which are not efficient.
The main mechanism in our protocols is to security compute the inner product. In 2009, Wong et.al. \cite{Wong2009} proposed a secure kNN (k-nearest neighbor) scheme on encrypted database, which developed a new asymmetric encryption that preserves inner product. We tailor the encryption scheme to meet our scenario, and construct our privacy preserving face retrieval application.

\textbf{Our Approach}. In this paper, we leverage a simple but efficient secure inner production protocol to protect the contents of user photos and the parameters of the face detector. The cloud server only provides the resources of storage and computing and can not learn anything of the user photos and the face detector. The face detection stage is protected by the secure face detection protocol using our secure inner production protocol. The face recognition stage is running locally in the users' mobile phone. The photos is encrypted and uploaded to the shared cloud storage space together with the corresponding label vector. The face label matching stage is running in the on-line face retrieval phase. The query label vector and the label vector in the cloud are compared using our secure inner production protocol. 

\begin{figure*}
\centering
\includegraphics[height=7.1cm]{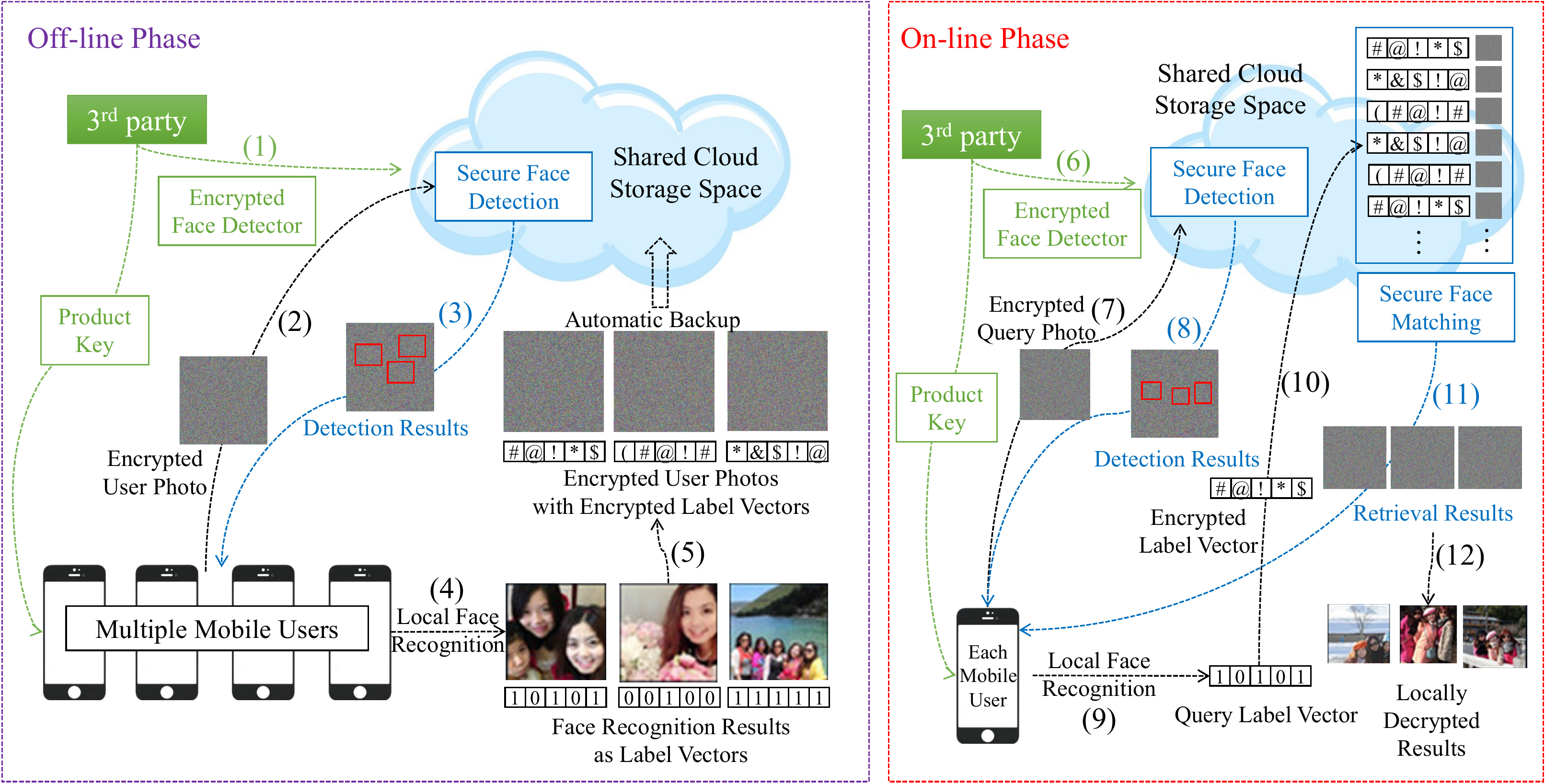}
\caption{The overall system architecture. (1) The face detector is provided by a 3rd party with a product key. (2) A user encrypts photos with the same key as used in (1), before sending them to the cloud. (3) Our secure face detector protocol is running in the cloud with the encrypted face detector and photos. The detected face windows are sent back to the user. (4) The local face recognition algorithm is called to mark each photo with a label vector, which reveals who is/are in this photo using $1$ for exist. (5) The photos in all the shared users are encrypted and uploaded to the cloud storage together with the encrypted label vectors. The off-line phase is end. In the on-line phase, a user want to query photos from the cloud storage with all the faces in the query photo. (6)(7)(8)(9) is the same as (1)(2)(3)(4). Then, the label vector of the query photo is computed. (10) The query label vector is encrypted and uploaded to the cloud and compared to all the label vectors in the cloud using our secure face matching  protocol. (11) The corresponding encrypted corresponding photos with the top $N$ matching label vectors are sent back to the user. (12) The user decrypts the matching photos and obtain the final retrieval results.}
\label{fig:overview}
\end{figure*}

\section{PROBLEM FORMULATION}
\subsection{Overview}
Our proposed methods are shown in Figure \ref{fig:overview}. In the face detection stage, face regions are detected in users' photos with a rectangle. In the recognition stage, each detected face are marked by a label of a member in a group. Then a label vector is generated according to the face recognition result for each photo so as to mark who is/are in each photo. The above is the off-line phase. In the on-line phase, a user queries a specific face of one person or faces of a group of person. Then, a label vector is generated for this query and compared to each label vector corresponding to each photo. Photos with the most similar label vector to the query label vector are selected as the retrieval result.

\subsection{Security Model}
We adopt the "honest-but-curious" model for the cloud server. It assumes that the cloud server would honestly follow the designated protocols and procedures to fulfill its service provider's role, while it may analyze the information stored and processed on the server in order to learn additional information about its customers. 

The objective of our scheme is to preserve the $3^{rd}$ party and users' data privacy, which includes: 1). face detector privacy; 2). detected windows privacy; 3). photos content privacy; 4). label vectors privacy; 5). query privacy. While photos content privacy can be achieved by encryption-before-outsourcing schemes, this paper focuses on preserving the data privacy due to the face detection and matching, as follows:
\begin{description}
\item[Detection Privacy] Besides the detection result, the cloud server should not deduce any face classifier information from the secure face detector, and face information from the secure detected windows. 
\item[Matching Privacy] Besides the matching result, the cloud server should not deduce any face information from the secure label vectors and secure query.
\end{description}

\section{Secure Face Retrieval}
\subsection{Secure Face Detection}

Denote some finite field $F$ that is large enough to represent all the intermediate results. Denote by $X$ the image that Alice owns. A particular detection window within the image $X$ will be denoted by $x \in F^L$ and $x$ will be treated in vector form. Bob owns a strong classifier of the form

\begin{eqnarray}
H(x) = \operatorname{sign}(\sum_{n=0}^{N-1}h_n(x)),
\label{eq:strong}
\end{eqnarray}
where $h_n(x)$ is a threshold function of the form

\begin{eqnarray}
h_n(x) = 
\begin{cases}
 & \alpha_n ~~  x^{T}y_n > \theta_n \\ 
 & \beta_n  ~~ \text{otherwise,}  
\end{cases}
\label{eq:weak}
\end{eqnarray}
and $y_n \in F^L$ is the hyperplane of the threshold function $h_n(x)$. The parameters $\alpha_n \in F, \beta_n \in F$ and $\theta_n \in F$ of $h_n(x)$ are determined during training; $N$ is the number of weak classifiers used.

As in Figure \ref{fig:overview}, step (1), the $3^{rd}$ party $3P$ first generate the product key according to the users $US$'s purchase as:
\begin{description}
\item[D-KeyGen(m):] Given a security parameter $m$ as the most length of the classifiers in the face detector, output the product key $SK(M_1,M_2,S)$, where $M_1,M_2\in\mathcal{R}^{m\times m}$ are randomly invertible matrices and $S\in\{0,1\}^m$ is a randomly vector. 
\end{description}
the next, $3P$ send this product key to $US$ via secure channel.

The second, $3P$ encrypt his classifiers in the face detector and upload to the cloud server with the detect parameter $\{\alpha_i,\beta_i,\theta_i\}_{i=1,\cdots,n}$.
\begin{description}
\item[E-FD(SK,Y):] To encrypt the classifiers $Y={y_1,\cdots,y_n}$ in the face detector, $3P$ split each vector $y_i$ into two vectors $\{y_i',y_i''\}$ following the rule: for each $y_{i,j}\in y_i$, set $y_{i,j}'=y_{i,j}''=y_{i,j}$ if $s_j\in S$ is $1$; otherwise $y_{i,j}'=\frac{1}{2}y_{i,j}-r$ and $y_{i,j}''=\frac{1}{2}y_{i,j}+r$ where $r\in\mathcal{R}$ is a random number. Then encrypt $\{y_i',y_i''\}$ with $(M_1,M_2)$ into $\{M_1^{T}y_i',M_2^{T}y_i''\}$. Output $EY=\{M_1^{T}y_i',M_2^{T}y_i''\}_{i=1,\cdots,n}$
\end{description}
As in Figure \ref{fig:overview}, step (2), to detect whether a detection window is a face, $US$ encrypt the window and upload to the cloud server.
\begin{description}
\item[E-DW(SK,w):] To encrypt the window $w$, $US$ split vector $w$ into two vectors $\{w',w''\}$ following the rule: for each $w_j\in w$, set $w_j'=w_j'=w_j$ if $s_j\in S$ is $0$; otherwise $w_j'=\frac{1}{2}w_j-r'$ and $w_j''=\frac{1}{2}w_j+r'$ where $r'\in\mathcal{R}$ is another random number. Then encrypt $\{w',w''\}$ with $(M_1,M_2)$ into $\{M_1^{-1}w',M_2^{-1}w''\}$. Output $EW=\{M_1^{-1}w',M_2^{-1}w''\}$
\end{description}
After receiving the secure classifiers and secure detected window, the cloud server output the detection results as in Figure \ref{fig:overview}, step (3).
\begin{description}
\item[DC(EY,EW):] For each secure classifier $\{M_1^{T}y_i',M_2^{T}y_i''\}$, the cloud server first compute 
$$t_i=(M_1^Ty_i')^T\cdot M_1^{-1}w'+(M_2^Ty_i'')^T\cdot M_2^{-1}w''=y_i^T\cdot w$$
\end{description}
and set $h_i=\alpha_i$ if $t_i\ge \theta_i$ or $h_i=\beta_i$ otherwise. At last, the could server output $H=sign(\sum_{i=1}^nh_i)$ as the detection result.

\subsection{Face Recognition and Label Vector}
After detect all the face in the photos, as in Figure \ref{fig:overview}, step (4), the users $US$ run the face recognition algorithm, i.e. SPR \cite{Wright2009}, and form the face label vector $L_i\in\{0,1\}^t$ for each photo, which describes who is/are in each photo. $US$ set the label set $\{L_i\}$ as the index of the photo.

\subsection{Secure Face Label Matching}
In order to build the secure label vectors of the photo set, the users $US$ first generate the private key as follows:
\begin{description}
\item[M-KeyGen(t):] Given a security parameter $t$ as the totally face number of the photo set, output the private key $PrK(N_1,N_2,T)$, where $S_1,S_2\in\mathcal{R}^{t\times t}$ are randomly invertible matrices and $T\in\{0,1\}^t$ is a randomly vector. 
\end{description}

As in Figure \ref{fig:overview}, step (5), $US$ encrypt his label vectors and upload to the cloud server.
\begin{description}
\item[E-LV(PrK,L):] To encrypt the label vector $L_i$ in the vector set, $US$ split each vector $L_i$ into two vectors $\{L_i',L_i''\}$ following the rule: for each $L_{i,j}\in L_i$, set $L_{i,j}'=L_{i,j}''=L_{i,j}$ if $t_j\in T$ is $1$; otherwise $L_{i,j}'=\frac{1}{2}L_{i,j}-u$ and $L_{i,j}''=\frac{1}{2}L_{i,j}+u$ where $u\in\mathcal{R}$ is a random number. Then encrypt $\{L_i',L_i''\}$ with $(N_1,N_2)$ into $\{N_1^{T}L_i',N_2^{T}L_i''\}$. Output $EL=\{N_1^{T}L_i',N_2^{T}L_i''\}_{i=1,\cdots,t}$
\end{description}
Next, The users $US$ choose the standard encrypted algorithm such as AES \cite{Daemen2002The}, or other photo encryption scheme such as \cite{Jin2015}, with their own secret key, to encryption the photos, and upload to the cloud server with the secure label vectors.

\begin{figure}
\centering
\includegraphics[height=2.8cm]{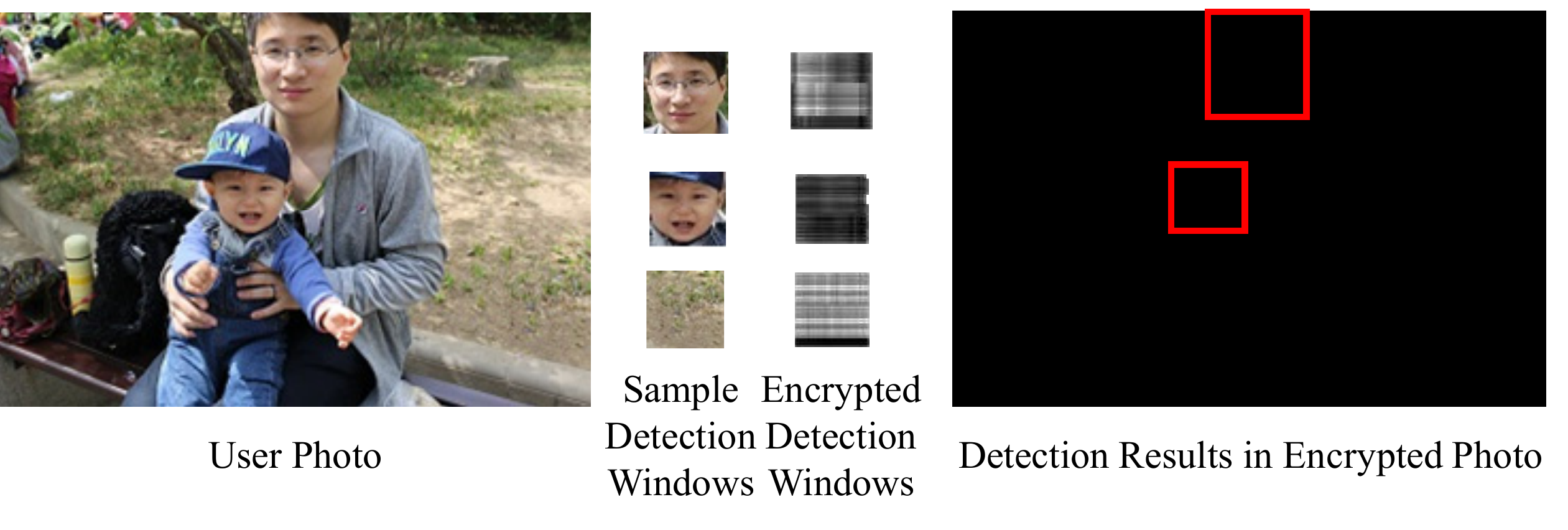}
\caption{The secure face detection. The user photo are divided into detection windows, which are sent to the cloud one by one. The detection results are shown in rectangle. }
\label{fig:facedetection}
\end{figure}

To search the photos with target faces, $US$ first generate the query as $Q=\{0,1\}^t$, as in Figure \ref{fig:overview}, step (9), where $Q_i=1$ if the i-th face is one of the target faces, then $US$ encrypt the query and upload it to the cloud server with the amount of target faces $\lambda$, as in Figure \ref{fig:overview}, step (10).
\begin{description}
\item[E-Q(PrK,Q):] To encrypt the query $Q$, $US$ split vector $Q$ into two vectors $\{Q',Q''\}$ following the rule: for each $Q_j\in Q$, set $Q_j'=Q_j'=Q_j$ if $t_j\in T$ is $0$; otherwise $Q_j'=\frac{1}{2}Q_j-v$ and $Q_j''=\frac{1}{2}w_j+v$ where $v\in\mathcal{R}$ is another random number. Then encrypt $\{Q',Q''\}$ with $(N_1,N_2)$ into $\{N_1^{-1}Q',N_2^{-1}Q''\}$. Output $EQ=\{N_1^{-1}Q',N_2^{-1}Q''\}$
\end{description}

After receiving the secure index and secure query, the cloud server output the matching result.
\begin{description}
\item[MAT(EL,EQ):] For each secure label $\{N_1^{T}L_i',N_2^{T}L_i''\}$, the cloud server first compute 
$$ret_i=(N_1^TL_i')^T\cdot N_1^{-1}Q'+(N_2^TL_i'')^T\cdot N_2^{-1}Q''=L_i^T\cdot Q$$
\end{description}
if $ret_i=\lambda$ the cloud server then set the $i-th$ photo with the label $L_i$ as one of the matching photos.

As in Figure \ref{fig:overview}, step (11), the cloud server return the encryption retrieval photos to $US$, then $US$ decryption them to get the matching photos.

\begin{figure}
\centering
\includegraphics[height=4cm]{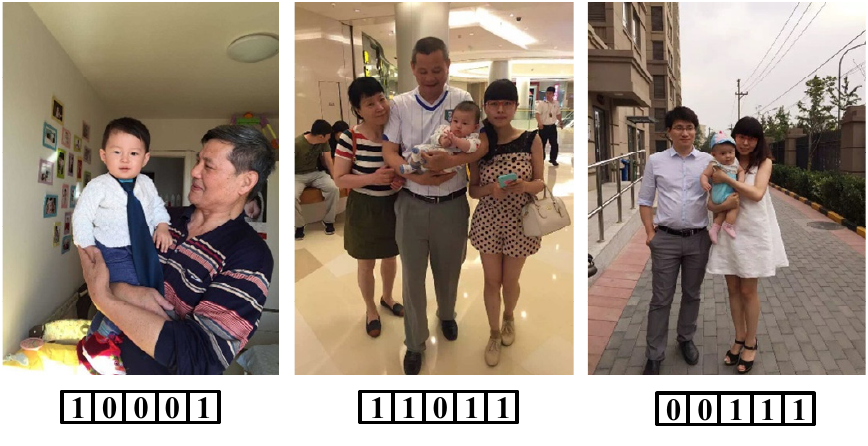}
\caption{The local face recognition. We label the recognition results in label vector, where 1 means that the corresponding person is in that photo. Then, the label vector is encrypted and sent to the cloud server. }
\label{fig:facerecognition}
\end{figure}

\section{Experiments}

We convert the Viola-Jones type face detector \cite{Viola2001,Viola2004} to our secure face detector, which is implemented by  OpenCV 2.4.3. \footnote{http://opencv.org/} package. The face detector consists of a cascade of 22 rejectors. The first rejector consists of 3 weak classifiers. The most complicated rejector consists of 213 weak classifiers. There is a total of 2135 weak classifiers. 


In this section, we show an experiment on photos from an authorized family, which consists of 5 family members with 4 mobile phones. We use 100 photos (20 photos for each member) to build the dictionary. The number of total family photos in the simulated cloud  is 1000. The secure face detection results are shown in Figure \ref{fig:facedetection}. The local face recognition results are shown in Figure \ref{fig:facerecognition}.
The secure face matching results are shown in the supplementary material.
 
%

\section{Conclusion}
In this paper, we propose a novel protocol to preserve the privacy of the users' photos and the parameters of the commercial face detector simultaneously in mobile cloud scenarios.  The experimental results reveal that our protocol can successfully retrieve the proper photos from the cloud server and protect the user photos and the face detector. 
One of the core of the convolutional neural network (CNN) is inner product. Thus, in the future work, we will extent our approach to privacy preserving deep learning framework for face retrieval.

\bibliographystyle{named}
\bibliography{ijcai17}

\end{document}